# Setting a Baseline for long-shot real-time Player and Ball detection in Soccer Videos

Konstantinos Moutselos*
*Dept. of Digital Systems*
*University of Piraeus*
Piraeus, Greece
kmouts@unipi.gr

Ilias Maglogiannis
Dept. of Digital Systems
*University of Piraeus*
Piraeus, Greece
imaglo@unipi.gr

*Abstract*—Players and ball detection are among the first required steps on a football analytics platform. Until recently, the existing open datasets on which the evaluations of most models were based, were not sufficient. In this work, we point out their weaknesses, and with the advent of the SoccerNet v3, we propose and deliver to the community an edited part of its dataset, in YOLO normalized annotation format for training and evaluation. The code of the methods and metrics are provided so that they can be used as a benchmark in future comparisons. The recent YOLO8n model proves better than FootAndBall in long-shot real-time detection of the ball and players on football fields.

## I. Introduction

Video Analytics attracts a lot of interest from the professional team sports industry, as it provides the basis for a wide range of applications, e.g., from supporting team management and tracking tactics to producing statistics for measuring players' performance or predicting the result and betting odds. To this end, there is a large development of related applications that process television images/videos and use the latest developments in deep learning tools for visual analysis. Developments in vision detection algorithms together with the development of powerful GPUs now provide the possibility to produce numerical data in real-time or near real-time. In any case, identifying the players and the ball and finding the camera's homography are the first steps in the football video processing pipelines [1].

We start as a baseline for the literature review from FootAndBall [2]. This work is referred to as one of the earlier attempts at player and ball recognition, with classic vision methods or with deep learning. The FootAndBall model was developed precisely for player and ball recognition from long-distance TV shots and set the standard in the field. Influenced by the YOLO [3] and SSD [4] object detectors, it was originally designed for fast execution during inference as a single-stage detector. It has a full-convolutional architecture that allows it to accept images of any resolution, even those different from the one it was trained on. To be able to locate objects as different in size as the ball to a player, it applies the Feature Pyramid Network technique [5]. It contains only 199k parameters and its publication reports on-par accuracy with the fine-tuned two-staged object detector Faster R-CNN [6] in player recognition and better accuracy in ball recognition. The code for the model architecture and training is provided, but the code for the evaluation and performance metrics for the players is missing. The datasets used for training and evaluation were the ISSIA-CNR Soccer [7] and Soccer Player Detection [8].

Since then, there are various publications, but they do not provide code on the proposed architectures or how to train and evaluate the results. In [9] an enhancement to the SSD [4] model is proposed for player detection, by adding a reverse-connected convolutional neural network (RC-CNN). Features extracted from higher layers are concatenated to lower convolutional layers. The reporting experiment resulted in advancement by ~1.8 mAP relative to SSD300. The training and evaluation are described to be conducted using the Soccer Player Detection [8] and the KITTI Pedestrian dataset.

The work from [10] ascertains the lack of public datasets for sports analysis, and regarding the player detection task the CenterTrack algorithm is utilized [11]. The reason given is that it showed better performance stability on two subsets from the ISSIA-CNR and Soccer Player Detection datasets, although FootAndBall and Faster R-CNN did better on one of the two datasets. CenterTrack for the object detection part, in turn, uses a center-point-based approach instead of bounding boxes: the CenterNet [11] architecture.

In a recent work [12], residual connections between the convolutional layers are utilized to enhance the players' detection performance. The datasets used were again the ISSIA-CNR and Soccer Player Detection, with random splitting 80/20 for training/testing. The reported AP performance analysis showed an advancement of 0.026 and 0.098 points on the two datasets compared to FootAndBall.

Lately, the 8th YOLO version was released by Ultralitics [13], with available open-source code. The YOLO architecture has gone through many changes since its first version in 2016 [3], such as the adoption or elimination of anchor boxes, loss functions, and various tricks and modifications, as the deep learning techniques evolved. The last published paper until now is describing the recent YOLO v7 [14]. Some concerns raised regarding the limitations of YOLO v7 are about its ability to detect small objects over different object scales and under varying light conditions [15]. However, these are exactly the difficulties in a football match, regarding the location of the ball and players. So, after the 8th version of the model, the question arises whether the latest version can favorably compete with FootAndBall in long-shot detections, where the ball is much smaller in size than the players, and doing so in a real-time mode.

So, next, we compare the performance of the two architectures FootAndBall and YOLO8 in player and ball detection. Along the way, we show that for their objective comparison, it is necessary to create an evaluation baseline. Such a baseline cannot be created without an appropriate dataset with reliable ground truth, common performance metrics, and common ways of evaluating performance. The advent of the recent SoccerNet v3 (https://www.soccer-net.org/) provides the solution for the data-related requirements of such a baseline. In this context, our work

This work was supported by the National Project DFVA – Deep Football Video Analytics– Hellenic General Secretariat of Research and Technology, [T2EΔK-04581], co-funded by the European Union - European Regional Development Fund.

involves the creation of such a dataset that we make available to the community.

Regarding the structure of the paper, after the Introduction follows: the Materials & Methods where we present the general workflow for the performance assessment and we describe the datasets of the work together with comments on the previous annotation inconsistencies, the DL environment, and architectures we used for the training and the comparison, together with the common performance metrics. In the Results, five models of the two architectures are evaluated on the same test bed. Following is the Discussion where the tackling of difficulties during the comparison of related models is presented, and last, the Conclusions of the work are provided.

## II. MATERIALS AND METHODS

Fig.1 shows the workflow used for the comparison of Player and Ball detection performance from long-shot football frames. First, using the recent SoccerNet v3 Dataset, we extract frames that correspond to distant shots and by processing its annotations we create the SoccerNet_v3_H250 dataset. As the dataset contains clearly defined segments for training/validation and test purposes, we use them as follows: We train from scratch the FootAndBall model (FBtr) and fine-tune the YOLO8n network with input resolutions 640x640 and 1200x1200 (Ytr640 & Ytr1200). In the training phase, we utilize the training part and the validation part for training performance monitoring.

After training, we compare on the SoccerNet_v3_H250 test-part a total of five models: the FootAndBall and YOLO8n models as available at publication (FBo and Yo640), and the other three (FBtr, Ytr640 & Ytr1200) that we trained with the SoccerNet_v3_H250 training/validation parts.
We use the same metrics and the same platform for training and their final assessment. The datasets with their annotation peculiarities, the models, and the ways of comparison are analyzed next.

### A. Soccer Video Datasets

*1) ISSIA-CNR [7]:* This dataset, as described in [2], consists of six synchronized Full-HD long-shot videos from the Italian league. The annotation includes 20,000 frames with player and ball bounding boxes. The annotations were produced by applying the algorithm proposed in [7] to facilitate the production of ground-truth annotations from video files in a semi-supervised manner. However, this algorithm introduces, in some cases, inaccurate detections into the ground truth. Fig. 2 shows an example of a frame from the ISSIA-CNR dataset. Gray arrows indicate cases where the bounding box contains only parts of the player's body. Because the algorithm tracks and predicts the movement of objects to minimize the need for manual annotation, when there is a sudden change in player movement, the position of the relevant box is often inaccurate.

The yellow arrow in Fig. 2 indicates the lack of ground truth of the goalkeeper, which came up during the image-preprocessing process from the FootAndBall code [16]. Because the preprocessing includes random cropping in the data augmentation methods, if the cropping coincides by even one pixel with the annotation box, this box is deleted from the list of player annotations. In the code we provide, we modified this function so that the box is deleted only if half of the box area is outside the cropped image.

The code provided by FootAndBall [16] includes all the functions for extracting the frames and annotations using the ".xgtf" XML format files. There, the existence of even more inconsistencies inherent in the ISSIA-CNR annotations, are mentioned in the source code [16], such as skipping the first 50 annotated frames due to wrong annotations, a note to use a smaller IoU ratio for mean precision evaluation in camera 5, existing shifted/reversed ball position annotations for some sequences, and the insertion of a delta variable due to frame number/ground truth discrepancy.

Another peculiarity of the dataset is that the position of the ball is not described in the same way as the players (bounding box), but using only x, and y coordinates for the center of the ball. This quirk affects several steps down the line, from how FootAndBall's architecture is designed, how prediction accuracy is assessed for the ball case, and how the ball ground-truth box is displayed on frames for visual verification. For the latter, to draw the ball annotation box shown in Fig. 2, an arbitrary determination is made in the size of the bounding box that encloses the center of the ball.

*2) Soccer Player Detection [8]:* This dataset consists of 2,019 images that correspond to two videos from professional football on the same field. The distribution of images in the two matches is 1,295 and 524 and the resolution is 1280x720. The players' bounding boxes have been manually annotated. However, there are also issues with ground truth. As shown in Fig. 3, there are cases where there is no player detection in phases where there is partial coverup or clipping. Also, annotations stop when a player or goalkeeper falls to the ground, and resume when they stand up again. Finally, there is no ball annotation.

*3) SoccerNet v3 [17]:* This is the recent and latest extension of a publicly available mass collection of 400 football matches. The matches have been split in a 290/55/55 ratio for training, validation, and testing, making it ideal for developing and comparing DL models. The number of frames is 24,459/4,797/4,730. Fig. 4 shows some dataset statistics regarding the training part: The left graph shows that there are five frame resolutions, although the great majority consists of 1280x720 and 1920x1080 resolutions. The central graph displays the distribution of the maximum (per frame)

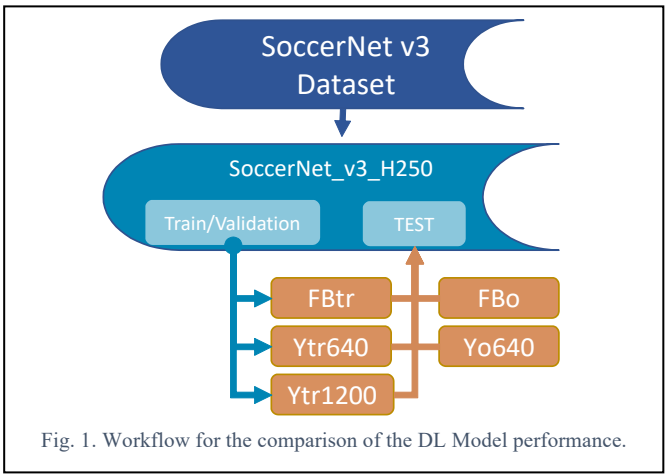

Fig. 1. Workflow for the comparison of the DL Model performance.

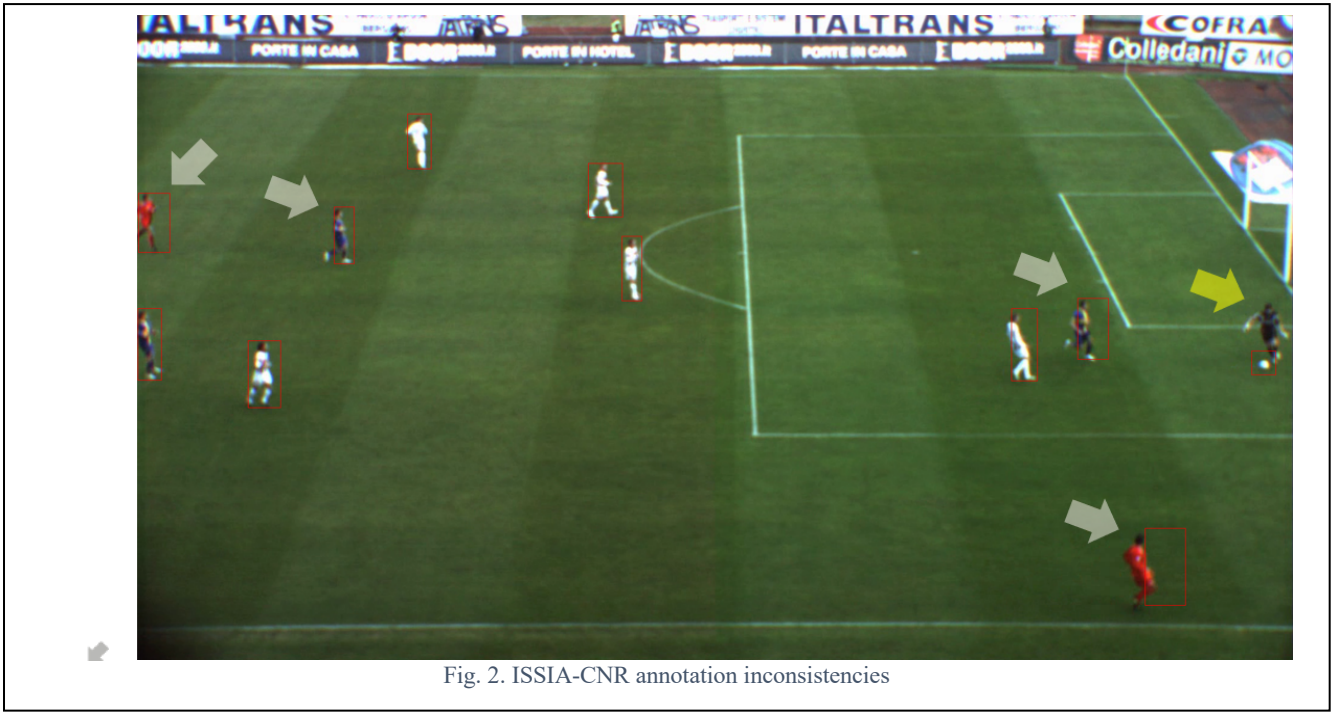
Fig. 2. ISSIA-CNR annotation inconsistencies

player height measured in pixels of the player bounding-box length. The right graph shows the distribution of the frames that include a ball. It should also be noted that the respective distributions of the graphs for the validation and testing tracks are similar, and thus no bias is introduced in the evaluation of the models. The only difference is that the validation and testing parts contain images with only 1280x720 and 1920x1080 resolutions.

The extensive and detailed annotations include - among others - bounding boxes for balls and humans. The human annotations are divided into 7 classes (left/right player/goalkeeper, main/side referee and staff). It should be noted that in the annotation of the dataset, special importance has been given to the interconnection of the frames of the "live" match with the corresponding replay action sequences. This approach reflects in the way the frames are annotated in .json files and processed by the relative data loader function. The SoccerNet v3 dataset is also escorted by source code [18] which greatly facilitates the loading and visualization of the data. With appropriate filtering of this dataset, results in the SoccerNet_v3_H250 dataset that we created in this work, together with the processed annotation files.

*4) SoccerNet_v3_H250:* This is our proposed dataset for evaluating long-shot player and ball detection models. It consists of the frames in which the length (height) of the bounding box that locates a person does not exceed 250 pixels. Also, the SoccerNet v3 compressed .png images have been extracted to the proper directory structure and converted to .jpg. Additionally, the corresponding annotations of the

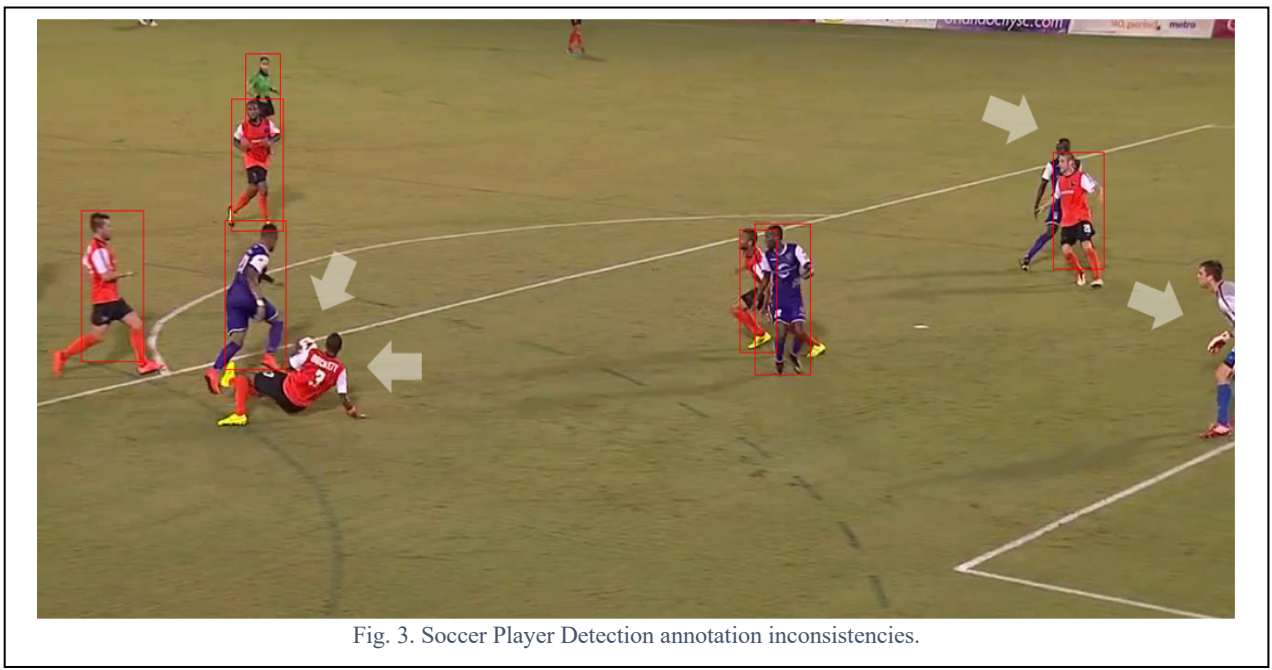
Fig. 3. Soccer Player Detection annotation inconsistencies.

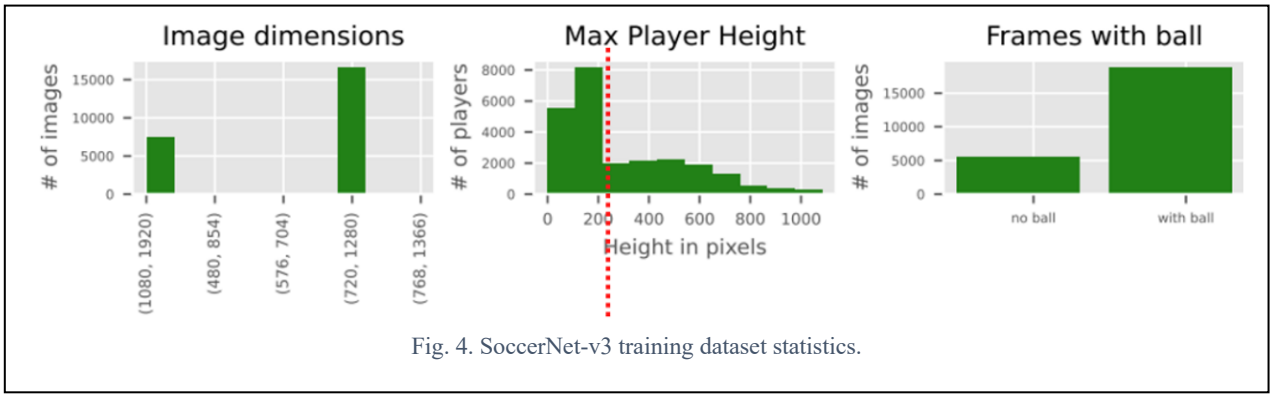

Fig. 4. SoccerNet-v3 training dataset statistics.

bounding boxes have been filtered and converted into a format compatible with YOLO annotation. Two classes are included: "0" for ball, and "1" for person bounding boxes. As "person" we have included all 7 classes of human annotations. Of course, the division into training, vali-dation and testing parts has been preserved, and the corresponding number of images is now: 14,368/2,726/2,692. In Fig. 4, the middle graph shows the threshold of 250 pixels in the Max_Player_Height distribution of images. The relevant code is also available: [https://github.com/kmouts/SoccerNet_v3_H250]

### B. DL Architectures

*1) FootAndBall*: As described in [2], the model consists of a series of convolutional blocks that reduce the size of the feature maps by a certain ratio. Subsequently, with the addition of FPA [5] and with parallel connections to merge features from different levels, two heads from a high level (player classifier and bounding-box regressor) are extracted, and only one head from a lower one (ball classifier). The choice of not having a ball-bounding box regressor is consistent with the particularity of the ISSIA-CNR dataset we mentioned above. Its exact number of trainable parameters is 198,840.

*2) YOLO8n[13]:* The latest version of YOLO by Ultralytics stresses the code rewriting effort and its' extensibility for model comparison. Although a paper has not been released yet, it is documented [19] that includes a new backbone network, an anchor-free detection head and a new loss function. We tested the nano version (8n) which is the smallest version of the model. Our model summary (fused) reported 168 layers and 3,006,038 parameters.

### C. DL environment, training and hyper-parameters

The sum of our experiments has taken place using an NVIDIA GeForce RTX 3080 GPU card, with Ubuntu OS, CUDA version 11.6, PyTorch [20] 1.13.1, and Python 3.8.15. The YOLO version is Ultralytics YOLOv8.0.39.

The pre-processing step for the FootAndBall evaluations (original and trained - FBo & FBtr in Table 1), involve the image normalization with the mean and standard deviation of the SoccerNet_v3_250 train dataset, without any image resizing. The FootAndBall training (Fbtr) parameters follow: epochs:200, lr:1e-3, batch_size:12. The augmentation operations on the training dataset before the same normalization included: CenterCrop to image size: 1280x720, ColorJitter with parameters bright-ness: [0.995, 1.005], contrast: [0.95, 1.05], saturation: [0.5, 1.5], hue: [-0.01, 0.01], and RandomAffine with parameters degrees:5, scale: [0.8, 1.2], p_hflip:0.5. On the validation dataset no augmentation operations are performed.

YOLO 8n 640 evaluations (original and trained - Yo640 & Ytr640 in Table 1) resize the input frames to 640x384 and then square to 640x640 for the inference.

YOLO 8n 1200 evaluation (trained - Ytr1200 in Table 1) resize the input frames to 1280x736 and then square to 1280x1280 for the inference.

For the YOLO 8n training (Ytr640 and Ytr1200 in Table 1), we opted for transfer learning (models pre-trained on the COCO dataset), which run with parameters: 100 epochs with patience 5, optimizer: SGD (lr=0.01), and imgsz: 640 and 1200 respectively. AutoBatch was used to determine the optimal batch size for each case: 53 and 14 for the two frame resolutions. For Ytr1200, the callback early-stopping was triggered, and the epoch-56 model was selected as the best model.

### D. Performance Metrics

We report our results on Players and Ball using both the $AP_{11}$ and COCO mAP Aver-age Precision metrics, utilizing the same torchmetrics MAP class [21]. The $AP_{11}$ for the Player class coincides with the "Player AP" reported in FootAndBall [2], which is a precision average for 11 (equally spaced) recall levels, with an Intersection Over Union - IOU threshold above 0.5. The other metric, COCO mAP is a more recently adopted and much stricter metric as it averages with IOU thresholds ranging between [0.5, …, 0.95] with step 0.05, and recall between [0, …, 1] with step 0.01.

Additionally, especially for the Ball class, we report the Average Precision (Avg. Prec. in Table 1) which coincides with the "Ball AP" metric reported in FootAndBall [2]. We utilized the same code as provided by FootAndBall [16], and we also report the Average Recall (Avg. Rec. in Table 1), and the percentage of correctly classified ball frames ("%" in Table 1). If TP, FP, and FN are the number of true positive, false positive, and false negative ball detections, then: Precision = TP / (TP + FP) and Re-call = TP / (TP + FN). We must stress that positive ball detection here means that the center of the predicted ball bounding box is within a 5-pixel radius from the center of the ground-truth bounding box. (The "Ball AP" metric reported in FootAndBall [2] is reportedly taken with a 3-pixel radius tolerance).

To estimate the real-time inference performance, we use the time in milliseconds (ms) required for inferencing ("I" in Table 1) of a frame. The YOLO8 code reports two metrics regarding time: the time just for inferencing and the overall time (along with image processing).

TABLE I. CLASSIFIER PERFORMANCE ON THE SOCCERNET V3 H250 DATASET

|  | Person | | Ball | | | | | I |
|---|---|---|---|---|---|---|---|---|
|  | $AP_{11}$ | COCO mAP | $AP_{11}$ | COCO mAP | Avg. Prec. | Avg. Rec. | % | ms |
| **FBo** | 0.3254 | 0.0771 | 0.0096 | 0.0003 | 0.783 | 0.022 | 0.234 | 9.5 |
| **FBtr** | 0.3905 | 0.1045 | 0.0165 | 0.0029 | 0.843 | 0.678 | 0.703 | 9.4 |
| **Yo640** | 0.7127 | 0.5195 | 0.1333 | 0.0370 | 0.524 | 0.118 | 0.284 | 7.2/9.0 |
| **Ytr640** | 0.9052 | 0.6824 | 0.3093 | 0.1207 | 0.856 | 0.410 | 0.518 | 7.3/9.2 |
| **Ytr1200** | **0.9058** | **0.7025** | **0.5361** | **0.2362** | **0.877** | **0.707** | **0.724** | 7.4/10.2 |

## III. RESULTS

Table 1 contains the results of our testing experiment. "FB" stands for the FootAndBall, and "Y" for the YOLO 8n models. The originally published models are denoted with "o", and our trained versions as "tr". The number next to the YOLO models stands for the input frame resolution that they support. Please refer to Section 2 for details on the metrics and reported models.

All the evaluations on the SoccerNet_v3_H250 Dataset have been performed averaging over the results with Batch size: 1. The best performances on accuracy metrics are denoted with bold font.

In the comparison of the results between FBo-FBtr we notice that the training in SoccerNet_v3_H250 offers a small improvement in the metrics related to player recognition, while in ball detection the improvement in the results is manifold. Especially in the recall of the ball - which is one of the most difficult tasks - the index rose from 2% to 68%. A similar trend is seen in the Yo640-Ytr640 comparison, where the ball metrics also show a significant improvement. Fine-tuning YOLO8n with higher resolution images, as seen in the Ytr640-Ytr1200 comparison, also show substantial advance for the ball case, while showing negligible improvement for the player case.

## IV. DISCUSSION

During the evaluation of the proposed in the literature state-of-the-art models for player and ball detection, several obstacles are presented. The first hurdle has to do with the annotated datasets used to train and evaluate the models. Some works use private datasets, so no equal comparison is possible. Until recently, open-access football datasets were either relatively small in size for training deep-learning models or had serious issues with ground-truth annotation, as described in the Materials and Methods section. But even in the case of two models using the same dataset, if the exact dataset split for training/validation/testing is not clear, this uncertainty still affects the fair comparison.

The second obstacle is related to the unavailability of the models training and evaluation code, so that the results could be verified. The dataset splitting/cross-validation procedures, as well as the image pre- and post-processing involved, matter in the evaluation, as well as the way the evaluation is done (e.g., with what batch size) also affect the results.

The third obstacle is the use of different metrics for evaluation. For example, the standard AP and mAP metrics change the way of calculation as the relevant vision contests evolve. The metric used, along with the associated code, and its version, should be fully specified.

Finally, another difficulty is the comparison of results when the testing is done on diverse hardware. The combination of these factors makes an objective comparison of the performance of the models practically impossible.

The recent advent of the SoccerNet v3 dataset can overcome the first hurdle. The filtering and further processing of this dataset introduced in this work kept only the long-shot shots. The threshold of 250 pixels in max player height per frame is not perfect, as we could utilize dynamic filtering considering the input frame resolution. Our approach, however, gives an easier and clearer treatment, and besides, the percentage of images that are smaller than 1280x720 is overwhelmingly smaller. When processing the annotations, as described in the Methods Section, we consolidated the human annotations into one class as "Persons". The rationale is that the first stage of a football analytics system is to identify all people near the field, and in a later stage, the separation from referees, staff, players outside the field, etc. The distribution of SoccerNet v3 into train/validation/test parts makes it much easier to deal with the second hurdle we mentioned earlier. Concerning addressing the third hurdle, we emphasize the importance of unambiguously indicating the performance metric. Here, to objectively compare the Ball and Players classes, we used AP in two specific versions, along with FootAndBall's original point-wise ball detection method.

The latest YOLO8n version is the clear winner in our comparison, and our interaction with the YOLO8 code and the available documentation, confirms the usability of the new version. It is also important, that although the number of parameters of the YOLO8n model is more than 15 times larger than FootAndBall, the difference in inference time is hardly more than 1 ms for the specific hardware. Additionally, YOLO 8 provides code with mix-precision optimization which could reduce even further its inference time. Here, we have not tested this feature of the code.

## V. CONCLUSIONS

In this work, we set a common baseline and compare two state-of-the-art models in their ability to detect players and balls in a football match, from distant camera shots. We focus on long-shot images because these frames can reveal the position of objects in the field and enable their projection (along with knowledge of the projection homography transform) into a 2D top view for the extraction of football statistics. The paper points out the inconsistencies and inadequacies of the datasets on which the majority of published architectures have relied so far. We propose and release to the community a processed SoccerNet v3 sub-dataset suitable for reliable evaluation of related models, as well as the code of the metrics we use so that it can be used as a benchmark in future comparisons. The Zenodo link for the proposed dataset is available at: [https://github.com/kmouts/SoccerNet_v3_H250] and the source code of the models and the metrics: [https://github.com/kmouts/FootAndBall]. Based on the proposed testbed, the YOLO8n model proves better than FootAndBall in long-shot detection of balls and players in real-time soccer fields.

REFERENCES

[1] P. Mavrogiannis and I. Maglogiannis, "Amateur football analytics using computer vision," *Neural Comput. Appl.*, Aug. 2022, doi: 10.1007/s00521-022-07692-6.

[2] J. Komorowski, G. Kurzejamski, and G. Sarwas, "FootAndBall: Integrated Player and Ball Detector," in *Proceedings of the 15th International Joint Conference on Computer Vision, Imaging and Computer Graphics Theory and Applications*, 2020, pp. 47–56. doi: 10.5220/0008916000470056.

[3] J. Redmon, S. Divvala, R. Girshick, and A. Farhadi, "You Only Look Once: Unified, Real-Time Object Detection." arXiv, May 09, 2016. Accessed: Mar. 16, 2023. [Online]. Available: http://arxiv.org/abs/1506.02640

[4] W. Liu *et al.*, "SSD: Single Shot MultiBox Detector," in *Computer Vision – ECCV 2016*, B. Leibe, J. Matas, N. Sebe, and M. Welling, Eds., in Lecture Notes in Computer Science. Cham: Springer International Publishing, 2016, pp. 21–37. doi: 10.1007/978-3-319-46448-0_2.

[5] T. Lin, P. Dollar, R. Girshick, K. He, B. Hariharan, and S. Belongie, "Feature Pyramid Networks for Object Detection," in *2017 IEEE Conference on Computer Vision and Pattern Recognition (CVPR)*, Jul. 2017, pp. 936–944. doi: 10.1109/CVPR.2017.106.

[6] S. Ren, K. He, R. Girshick, and J. Sun, "Faster R-CNN: Towards Real-Time Object Detection with Region Proposal Networks," *ArXiv150601497 Cs*, Jun. 2015, [Online]. Available: http://arxiv.org/abs/1506.01497

[7] T. D'Orazio, M. Leo, N. Mosca, P. Spagnolo, and P. L. Mazzeo, "A Semi-automatic System for Ground Truth Generation of Soccer Video Sequences," in *Proceedings of the 2009 Sixth IEEE International Conference on Advanced Video and Signal Based Surveillance*, in AVSS '09. USA: IEEE Computer Society, Sep. 2009, pp. 559–564. doi: 10.1109/AVSS.2009.69.

[8] K. Lu, J. Chen, J. Little, and H. He, "Light Cascaded Convolutional Neural Networks for Accurate Player Detection," in *Procedings of the British Machine Vision Conference 2017*, London, UK: British Machine Vision Association, 2017, p. 173. doi: 10.5244/C.31.173.

[9] L. Zhang, Y. Lu, G. Song, and H. Zheng, "RC-CNN: Reverse Connected Convolutional Neural Network for Accurate Player Detection," in *PRICAI 2018: Trends in Artificial Intelligence*, X. Geng and B.-H. Kang, Eds., in Lecture Notes in Computer Science. Cham: Springer International Publishing, 2018, pp. 438–446. doi: 10.1007/978-3-319-97310-4_50.

[10] J. Theiner, W. Gritz, E. Müller-Budack, R. Rein, D. Memmert, and R. Ewerth, "Extraction of Positional Player Data from Broadcast Soccer Videos," presented at the 2022 IEEE/CVF Winter Conference on Applications of Computer Vision (WACV), IEEE Computer Society, Jan. 2022, pp. 1463–1473. doi: 10.1109/WACV51458.2022.00153.

[11] X. Zhou, V. Koltun, and P. Krähenbühl, "Tracking Objects as Points," in *Computer Vision – ECCV 2020: 16th European Conference, Glasgow, UK, August 23–28, 2020, Proceedings, Part IV*, Berlin, Heidelberg: Springer-Verlag, Aug. 2020, pp. 474–490. doi: 10.1007/978-3-030-58548-8_28.

[12] T. Wang and T. Li, "Deep Learning-Based Football Player Detection in Videos," *Comput. Intell. Neurosci.*, vol. 2022, p. 3540642, Jul. 2022, doi: 10.1155/2022/3540642.

[13] G. Jocher, A. Chaurasia, and J. Qiu, "YOLO by Ultralytics." Jan. 2023. Accessed: Mar. 19, 2023. [Online]. Available: https://github.com/ultralytics/ultralytics

[14] C.-Y. Wang, A. Bochkovskiy, and H.-Y. M. Liao, "YOLOv7: Trainable bag-of-freebies sets new state-of-the-art for real-time object detectors." arXiv, Jul. 06, 2022. doi: 10.48550/arXiv.2207.02696.

[15] R. Kundu, "YOLO Algorithm for Object Detection Explained." https://www.v7labs.com/blog/yolo-object-detection, https://www.v7labs.com/blog/yolo-object-detection (accessed Mar. 19, 2023).

[16] "GitHub - jac99/FootAndBall: FootAndBall: Integrated player and ball detector," *GitHub*. https://github.com/jac99/FootAndBall (accessed Mar. 17, 2023).

[17] A. Cioppa, A. Deliège, S. Giancola, B. Ghanem, and M. Van Droogenbroeck, "Scaling up SoccerNet with multi-view spatial localization and re-identification," *Sci. Data*, vol. 9, no. 1, Art. no. 1, Jun. 2022, doi: 10.1038/s41597-022-01469-1.

[18] "SoccerNet-v3." SoccerNet, Mar. 14, 2023. Accessed: Mar. 18, 2023. [Online]. Available: https://github.com/SoccerNet/SoccerNet-v3

[19] "YOLOv8 Docs." https://docs.ultralytics.com/#ultralytics-yolov8 (accessed Mar. 19, 2023).

[20] A. Paszke *et al.*, "PyTorch: An Imperative Style, High-Performance Deep Learning Library," in *Advances in Neural Information Processing Systems*, Curran Associates, Inc., 2019. Accessed: Mar. 19, 2023. [Online]. Available: https://papers.nips.cc/paper/2019/hash/bdbca288fee7f92f2bfa9f7012727740-Abstract.html

[21] "Mean-Average-Precision (mAP) — PyTorch-Metrics 0.11.4 documentation." https://torchmetrics.readthedocs.io/en/stable/detection/mean_average_precision.html (accessed Mar. 19, 2023).